\documentclass[runningheads]{llncs}
\usepackage[T1]{fontenc}
\usepackage{graphicx}
\usepackage{amsmath}
\usepackage{amssymb}
\usepackage{mathabx}
\usepackage{cite}
\usepackage{algorithm}
\usepackage{algpseudocode}
\usepackage{hyperref}
\usepackage{color}

\begin{document}

\title{Extracting Semantics: LLM-Guided Automatic Population of Robot Ontology from URDF}
\titlerunning{Semantic Robot Body-Awareness}
%
\author{Bastien Dussard\orcidID{0009-0001-8800-8853} \and
Guillaume Sarthou\orcidID{0000-0002-4438-2763}}
\authorrunning{B. Dussard \and G. Sarthou}
%
\institute{LAAS-CNRS, Department of Robotics, Toulouse, France
\email{firstname.surname@laas.fr}}
\maketitle              
\begin{abstract}

While commonsense knowledge may suffice for virtual agents, embodied robots interacting with humans require grounded and semantically rich representations of both their environment and their own physical embodiment. In cognitive robotics, ontologies are effective for integrating such heterogeneous knowledge to enable explainable reasoning, even during continuous knowledge updates. Yet, their manual construction remains a bottleneck. We present a preliminary approach for the automatic generation of robot semantic abstractions by transforming Unified Robot Description Format (URDF) models into populated ontologies. Although URDF files provide structural and kinematic descriptions, their identifiers often require commonsense interpretation to recover meaningful semantics, a task at which Large Language Models (LLMs) excel. Our pipeline leverages LLMs to infer semantic relationships by prompting them with concepts from an existing ontology, ensuring the final classification remains aligned with the formal model.
To improve reliability, the pipeline combines majority voting across multiple LLM queries along with syntactic and schema-level validation to ensure that generated outputs conform to the expected representation format and ontology constraints. We evaluate the approach on multiple robot descriptions and discuss the generated abstractions. Initial results indicate that the proposed method can effectively bridge the gap between low-level robot descriptions and the structured, grounded knowledge representations required for human–robot interaction.

\end{abstract}
\section{Introduction}

For robots to act effectively in the physical world, commonsense knowledge is mandatory but not sufficient as they also require grounded knowledge of both their surroundings and their own embodiment. Indeed, knowing that a bottle can water a plant is a useful start. True autonomy, however, requires understanding which specific containers are present and which ones the robot can physically manipulate. In the context of Human-Robot Interaction (HRI), communicating this grounded knowledge is essential: it allows the robot to parse human instructions ("Bring me a container"), provide informative feedback ("I cannot, as containers are stored upstairs and I cannot climb stairs with my wheels"), and learn new skills through demonstration ("If your hands are full, you can put the towel on your forearm"). In such scenarios, a robot requires multiple interconnected levels of knowledge: commonsense knowledge (e.g., what an "arm" is), grounded knowledge (e.g., "Do I have arms, and which specific joints constitute them?"), and technical knowledge (e.g., the specific joint limits of those arms). Capturing and structuring this multi-level knowledge in a machine-readable form is essential for autonomous reasoning and explainable HRI.

While the integration of commonsense knowledge in cognitive robotics has been a bottleneck for decades~\cite{toberg2024commonsense}, the rise of Large Language Models (LLMs) offers new possibilities. However, despite their high performance on diverse tasks, LLMs have inherent limitations regarding grounded, robot-specific knowledge. Specifically, they may suggest plans that are physically unfeasible given a robot's specific configuration~\cite{brohan2023can} or lack the multi-modal physical grounding necessary to perceive the robot’s immediate environment accurately~\cite{driess2023palm}. Aligning an LLM with a particular robot embodiment and its capabilities requires either fine-tuning or providing knowledge via prompting. Fine-tuning lacks generality and adaptability, whereas prompting depends on externally provided grounded knowledge. Moreover, a robotic agent relying solely on LLMs often lacks a centralized, persistent worldview and may require repeated, inefficient queries for static information that evolves slowly over time. To bridge this gap, grounded knowledge should be maintained independently in structured resources that can be queried, verified, extended, and reasoned upon, providing high-quality, contextually grounded input to the LLM.

In the field of knowledge representation and reasoning applied to robotics, ontologies have gained increasing attention as a means to improve robots' autonomy~\cite{olivares2019review}, supporting both task planning~\cite{beetz2018know} and interactive communication strategies~\cite{gomez2021ontology}. Highly structured, ontologies enable verification and inference through the use of formal logic semantics (i.e., first-order or description logic). Their semantic nature provides rich abstractions through class and property hierarchies, bridging the gap between commonsense concepts (e.g., the $Camera$ class) and grounded knowledge stored in instances (e.g., $camera_1$ as an instance of $Camera$). Some frameworks further support dynamic ontology usage, making them suitable for online learning. However, constructing these knowledge bases remains labor-intensive, requiring substantial manual effort from domain experts and limiting widespread adoption. To overcome this challenge, these ontologies can leverage the commonsense reasoning capabilities of LLMs to automate the population process, reducing manual effort while maintaining structured, semantically rich knowledge.

While contemporary research often leverages LLMs to generate ontologies directly~\cite{zhu2024llms}, our work adopts a different paradigm: using LLMs as a classification heuristic within a formal, verifiable framework. We focus on representing robot self-embodiment. Although most modern robots encode embodiment using the Unified Robot Description Format (URDF), this format lacks standardized semantic information. Existing approaches, such as the Semantic Robot Description Format (SRDF)~\cite{chitta2012moveit} or Semantic Robot Description Language (SRDL)~\cite{kunze2011towards}, provide formal semantics but still require extensive manual annotation.

Our approach is motivated by the observation that human experts can infer functional characteristics, such as locomotion type, sensor assemblies, and end-effector configurations, by interpreting the semantically rich (but informal) identifiers in URDF files. By treating URDF as a semi-structured source of grounded knowledge, LLMs can emulate this expert reasoning, bridging raw structural data and formal semantic abstractions. URDF is therefore well-suited for LLM-guided ontology population, enabling structured, grounded knowledge for complex human-robot interaction without extensive manual modeling.

The primary contribution of this work is a partially closed-loop framework to automatically populate ontologies from semi-structured URDF data using LLMs. Unlike open-loop approaches that treat LLMs as black-box generators, our method constrains the model by prompting it with specific concepts from the target ontology, ensuring strict alignment. To mitigate hallucinations, we implement a robust verification pipeline combining multi-stage classification, output validation, and majority voting, effectively filtering irrelevant outputs and preserving ontological integrity. We demonstrate the framework's efficacy on sensors and end-effectors, establishing a proof-of-concept for broader embodiment extraction. While comprehensive ontology modeling remains complex, these preliminary results validate our pipeline as a scalable foundation for bridging raw robot descriptions and structured, machine-reasoning-ready knowledge.

\section{Related Work}

The integration of commonsense knowledge into knowledge-based systems is a prerequisite for high-level robotic autonomy. This challenge falls within the broader scope of ontology creation, which can be divided into two primary tasks: Ontology Learning (OL), which extracts conceptual schemas and terminology, and Ontology Population (OP), which asserts individuals into existing terminology. Traditionally, these tasks have relied on human-in-the-loop methods but the emergence of LLMs offers a new mechanism for extracting implicit semantic patterns from heterogeneous data sources~\cite{remadi2024prompt}.

\paragraph{LLMs in Ontology Engineering: From Text to Structure.}

Recent research has explored using LLMs to support semi-automated pipelines for Ontology Learning (OL) and Ontology Population (OP). For instance, Kommineni et al.~\cite{kommineni2024human} use LLMs to generate competency questions (CQs) about the ontology domain. These CQs are then employed to automatically generate the required taxonomy and populate it from textual corpora. Similarly, Funk et al.~\cite{funk2023ontology} leverage LLMs as generators of conceptual taxonomies, though without population. While these methods effectively transform unstructured natural language into formal representations, they have a key limitation for robotics: they rely on the LLM as the primary source of knowledge, whereas robots require LLMs to act as interpreters of grounded data, such as URDFs.

Some approaches focus on OP from existing ontologies. In some cases, the LLM is provided with the entire ontology specification~\cite{sahbi2024automatic}, while others prompt the model with relevant Ontology Design Patterns (ODPs)~\cite{shyama2025pattern} or schema relations~\cite{norouzi2024ontology} over textual corpora. However, these methods typically operate in an open-loop setting, where the LLM generates outputs without constraints from the evolving formal ontology. This can lead to semantic drift or hallucinations, limiting the reliability of the populated knowledge base.

\paragraph{Commonsense Reasoning in Robotic Ontologies.}

In the robotics domain, similar techniques have been explored to leverage the commonsense knowledge encoded in LLMs. Adamik et al.~\cite{adamik2025extracting} demonstrate that even small-scale LLMs can reliably suggest possible properties for everyday objects, which can later be used for ontology population. In the context of HRI, Nakajima et al.~\cite{nakajima2024combining} employ LLM-based commonsense reasoning to disambiguate commands, using an ontology as a post-generation validation step.
Despite these successes, these frameworks predominantly treat LLMs as black-box generators. Even when Ontology Design Patterns guide the model (such as Ocker et al.~\cite{ocker2023exploring}), the extraction process remains unidirectional. In contrast, our work employs the LLM not as a knowledge repository but as a heuristic reasoning component operating under the strict constraints of a predefined ontological schema, ensuring that extracted embodiment data is structured and reasoning-ready.

\paragraph{Semantic Refinement of Robot Descriptions.}

For robot morphology, the Semantic Robot Description Format (SRDF)~\cite{chitta2012moveit} and Semantic Robot Description Language (SRDL)~\cite{kunze2011towards} represent foundational attempts to add meaning to raw structural data. SRDF enables the grouping of kinematic chains but lacks a formal semantic language, limiting its applicability in higher-level ontological reasoning. SRDL provides a more formal integration by importing URDF links into an ontology but remains heavily dependent on manual expert refinement or specific ROS-based tags.
A key bottleneck in these systems is the semantic interpretation of non-standardized identifiers (e.g., $L_arm_j1$). Our approach addresses this challenge by closing the loop between structural URDF data and ontologies. By introducing a multi-stage classification pipeline with majority voting and output verification, we automate the "expert task" of semantic abstraction. This transforms the URDF from a purely kinematic description into a structured, semantically grounded representation suitable for reasoning and downstream ontology population.

\section{Methodology}
The proposed pipeline transforms a robot's structural URDF model into a semantically grounded ontology by resolving the naming ambiguities inherent in the source file. Preliminary experiments showed that baseline approaches treating the URDF as a flat textual corpus for competency-question-driven ontology population produce unreliable results because they ignore the hierarchical and kinematic structure encoded in the URDF.

To address these structural requirements, our pipeline combines deterministic URDF analysis with the semantic inference capabilities of LLMs, ensuring that extracted components are both semantically interpreted and aligned with the target ontology terminology. The workflow incorporates built-in verification mechanisms to preserve ontology integrity: multiple queries to the LLM~\footnote{The model is queried with a relatively high temperature of 0.7 to obtain generation variability.} are aggregated via majority voting to reduce variability, syntactic validation ensures outputs reference valid URDF identifiers and adhere to the expected format, and schema-level validation constrains labels to a predefined set of ontology concepts.

The remainder of this section is organized as follows. We first provide a brief overview of the URDF standard, followed by a detailed description of the seven stages of the pipeline: Identification, Attachment Root Detection, Pre-Classification, Group Merging, Post-Classification, Abstraction Verification, and Ontology Population.

\subsection{Unified Robot Description Format}

A Unified Robot Description Format (URDF) model represents the hierarchical structure of a robot as a graph in which links correspond to nodes and joints correspond to edges. Each link and joint is assigned a unique identifier, while joints are further characterized by their actuation type (e.g. fixed, revolute, prismatic). For instance, a simple URDF chain element may specify that the link $linkA$ is connected to the link $linkB$ through a fixed joint named $jointAB$. Although URDF files can be extended with additional information through plugins (e.g. those used in Gazebo), this work considers only the minimal set of information required in a valid URDF model. The only available information are thus the names of links and joints, the joints type, and the structure.

As URDF initially tackles technical requirement, one can notice that a single physical component may be represented by multiple links in the model, reflecting the presence of sub-assembly frames. For example, a camera may include several optical frames corresponding to different lenses, as well as additional sensing elements such as an IMU. Consequently, the objective of our pipeline is to identify the various sub-parts associated with a given component, classify them according to our taxonomy, and subsequently regroup them into a coherent component assembly. This assembly can then be further classified as a higher-level component type, which will be later be abstracted.

\subsection{Pipeline Overview}

The following pipeline iterates over different types of components to abstract from the URDF model. In this work, we focus first on sensors and then on end-effectors, although we believe that the approach could be extended to body parts. For clarity, all examples that follow focus on sensors but the pipeline is strictly the same for both sensors and end-effectors.

\paragraph{\textbf{Step 1: Identification Task.}}
For a given component type (e.g., sensors), the first stage of the pipeline identifies links potentially related to that type. As a naive approach, we tested classification via competency questions over the entire URDF and the full link list. As expected, this often missed relevant links or incorrectly included others, since even a single component can involve many links. To reduce reasoning complexity and ensure token efficiency, we limit the search space to leaf links, which in typical URDFs correspond to sensors and end-effectors at kinematic chain extremities. The LLM is then tasked with selecting only the links relevant to the component type (e.g., $sensor\_link$ would be selected, $gripper\_link$ discarded). At this stage, the exact type of sensor is unknown, but all candidate sensor links are captured.

\paragraph{\textbf{Step 2: Attachment Root Detection Task.}}
Once the relevant component leaf links are identified, the pipeline focuses exclusively on these links, avoiding clutter from irrelevant information. To abstract the component from its upper structural parts, it is necessary to detect the component's attachment root. Leaf links alone (e.g., a camera's optical frame) may miss critical links, such as the camera's base or internal frames. This stage provides an initial estimate of the component’s structural boundary, reducing complexity for subsequent steps. The LLM is tasked with identifying, for each leaf link, its attachment root along the kinematic chain up to the URDF root, representing the link from which the component can be detached.

\paragraph{\textbf{Step 3: Pre-Classification Task.}}
Once relevant sub-chains are isolated, the pipeline determines their functional types by mapping them to formal ontology concepts. To ensure semantic alignment and reduce LLM hallucinations \cite{sahbi2024automatic}, we adopt a constrained, iterative classification strategy inspired by \cite{funk2023ontology}. Rather than performing open-ended labeling, the LLM is guided by a dynamic subset of candidate concepts obtained from the ontology at each taxonomy level. For each trimmed sub-chain (from attachment root to leaf), the model selects the most appropriate parent concept, triggering retrieval of more specific sub-concepts in subsequent iterations. This recursive process continues until a "leaf concept" is reached or classification converges, ensuring valid ontology membership and enabling finer-grained abstraction of internal frames to support the following grouping task.

\paragraph{\textbf{Step 4: Group Merging Task.}}
Once the sub-components have been classified, it is necessary to account for potential sub-assemblies among them. One option would be to provide the LLM only with already processed chains of links, for example grouping links attached to the same attachment root. However, this approach could overlook links that are attached directly to a parent link without a shared child, even if they belong to the same component group. In addition, it could group distinct components which are connected to the same parent link. To address this, we provide the LLM with all classified sub-chains up to their attachment roots and prompt it to identify coherent component groups. For instance, the LLM could output a group consisting of two leaf links that would represent a camera's RGB and depth frames. As a result of this step, we obtain groups of links corresponding to real-world physical devices, which can then be re-classified with greater precision.

\paragraph{\textbf{Step 5: Post-Classification Task.}}
Since component groups may include multiple sub-components, higher-level types can be inferred from their prior classifications. For instance, a camera group containing both RGB and depth sub-components can be reclassified as a single RGBD camera. To achieve this, we perform a post-classification step analogous to the iterative pre-classification, but applied to component groups rather than individual links. Each group, along with its sub-chains, attachment roots, and previous classifications, is provided to the LLM for refinement. The LLM is tasked with identifying a coherent parent concept that best represents the combined functionality of the sub-components, ensuring semantic consistency with the ontology. 

\paragraph{\textbf{Step 6: Abstraction Verification Task.}}
After having identified our sub-components, located their mounting points, classified them, grouped them into coherent ensembles, and refined their upper classification, we now have the necessary information to abstract these groups into functional entities. Before proceeding, we must ensure that removing these groups from the URDF does not omit any unclassified sub-links that may belong to other components. Starting the pipeline with sensor components (which are typically located at the leaves of kinematic chains) helps mitigate this risk, but it cannot be assumed in all cases. Therefore, we verify that none of the group's elements has a child node outside the group.

\paragraph{\textbf{Step 7: Ontology Population Task.}}
After executing the pipeline for a given component type, the identified elements are used to construct an abstraction of the corresponding components, which are then removed from the URDF to enable further stages. Each component is represented by an identifier, a mounting point, an upper-level concept, and optionally lower-level concepts for its sub-links.
To populate the ontology based on those functional representation, a new individual is created for the component group (e.g., $sensor_1$) and asserted as an instance of its post-classification type (e.g., $RGBDCamera(sensor_1)$). Then, lower-level components are similarly created and asserted with their pre-classification (e.g., $RGBCamera(subsensor_{11})$, $DepthCamera(subsensor_{12})$). Finally, object properties are asserted to capture the assembly structure between parent and child links (e.g., $hasComponent(sensor_1, subsensor_{11})$).

To facilitate the task for subsequent component abstraction pass, the sub-trees of abstracted components are replaced with a single node attached to their identified attachment root link (e.g., $mount\_link \rightarrow sensor_1$). Indeed, since the pipeline's usage is iterative, each pass reduces the search space for subsequent stages. For example, after sensor abstraction, all sensor assemblies are condensed into single nodes, enabling the LLM to focus more on relevant links during the subsequent end-effector stage.

\section{Preliminary Results}

To evaluate the proposed pipeline, we selected a LLM capable of on-robot deployment within a 16GB VRAM hardware envelope ($gpt\!-\!oss\!:\!20b$). This model is a Mixture-of-Experts (MoE) architecture that incorporates a native reasoning mode. It allows the model to perform complex, multi-step deliberation by generating an internal chain-of-thought prior to producing the final answer. While this study employs this model, the pipeline is designed to be model-agnostic and compatible with other LLM architecture, regardless of their reasoning capability or not. Our pipeline employs a partially-closed loop mechanism that leverages an ontology we developed as domain experts to ground the classification of sensor and end-effector components. By utilizing this formal knowledge base as a constraint for iterative classification, the system ensures ontological integrity, forcing the LLM to align its structural reasoning with standardized robotic concepts rather than generating arbitrary categories. 

For this preliminary study, we focused on socially-situated environments, selecting humanoid and service-oriented platforms as our primary benchmarks. Unlike traditional industrial manipulators characterized by rigid, repetitive serial chains, these robots exhibit high morphological complexity, featuring dense, multi-modal sensor arrays and articulate end-effector assemblies. Consequently, we validated our abstraction process across three diverse platforms: the PR2, Pepper, and TIAGo robots.

The preliminary results, including the used ontology, source code and experimental data, are hosted on a dedicated online repository\footnote{Link to the GitHub repository: \hyperref[https://github.com/RIS-WITH/urdf2ontology]{https://github.com/RIS-WITH/urdf2ontology}}. In the following sections, we provide a concise description of each robot's physical configuration followed by an analysis of the resulting ontological abstraction produced by our pipeline.

\subsection{TIAgo evaluation}
The TIAGo robot features a modular architecture comprising a mobile base, which is equipped with a front-facing 2D laser scanner, an internal IMU, and three rear ultrasonic sonar sensors. Its sensor head, mounted on a lifting torso, contains an RGB-D camera for 3D perception and a stereo microphone array. The chosen model's end-effector is a parallel gripper.

\paragraph{Sensor:} The abstraction of TIAGo’s sensor suite successfully identified and partitioned 10 distinct functional units, demonstrating the pipeline's ability to handle spatially distributed exteroceptive sensors. The mobile base assembly features a high degree of modularity where the $Lidar$, $IMU$, and three separate $Sonar$ units were correctly isolated as individual components attached to the $base\_link$ root. While the microphone links were successfully identified, they were not clustered into a single microphone array component. This behavior can be attributed to the URDF topology because each microphone is mounted directly to the root link, without a dedicated common mounting link. As a result, the LLM preserves their independence as separate $Microphone$ components. If these links were nested within a shared structural mount, the pipeline would likely have aggregated them into a unified group. On the other hand, in the upper body, the complex head-mounted perception system was effectively clustered. The pipeline successfully aggregated the color, depth, and general optical frames into a single $RGBD\_Camera$ group, resolving the multi-step kinematic subchain through intermediate frames.

\paragraph{End-Effector:} For the manipulation system, the pipeline successfully identified the \textit{ParallelGripper} assembly, trimming the kinematic chain at the \textit{wrist\_ft\_tool\_\\link} attachment point. The grouping task correctly clustered the physical and operational elements of the end-effector into a single coherent unit. The actuated components, specifically the \textit{gripper\_left\_finger\_link} and \textit{gripper\_right\_finger\_\\link}, were accurately pre-classified as \textit{ContactSurface} sub-parts, identifying their role in object interaction. However, the abstraction of virtual and reference frames revealed a consistent semantic trend: the \textit{gripper\_tool\_link} was categorized as a \textit{ToolHousing}, while the \textit{gripper\_grasping\_frame} was labeled a \textit{StructuralSubpart}. Although these virtual frames would be more precisely defined as reference origins or tool center points, the grouping logic's reliance on the shared \textit{gripper\_link} parent ensured that these classification nuances did not lead to fragmentation. In fact, this misclassification is not an inherent failure of the pipeline, but a consequence of the taxonomic hierarchy. When semantically similar concepts exist at a shared depth, a "branching effect" occurs. Once the LLM selects a specific branch, it is limited to the sub-concepts of that lineage, effectively propagating the initial error downward.

\subsection{PR2 evaluation} 

The PR2 robot is characterized by a dual-arm configuration and an extensive sensor suite designed for mobile manipulation. Its head assembly houses a tilting laser scanner, a Prosilica color camera, wide and narrow stereo camera systems, and a RGB-D Kinect camera, while the base integrates a primary 2D Lidar and an internal IMU. Each of its two arms terminates in a parallel-jaw gripper equipped with pressure-sensitive fingertip tactile arrays, an accelerometer and a dedicated forearm camera.

\paragraph{Sensor:} The PR2 robot presented a significantly higher degree of topological complexity due to its dense, nested sensor head and peripheral sensing capabilities. The abstraction successfully partitioned 12 distinct sensor units, demonstrating high precision in resolving complex kinematic branches. In the head assembly, the pipeline successfully differentiated between the narrow and wide stereo systems. Despite sharing a common ancestor ($double\_stereo\_link$), they were correctly clustered into two separate $StereoCamera$ components. Similarly, the $head\_mount\_kinect$ was accurately aggregated into an $RGBD\_Camera$ unit by resolving the relationship between its infrared and RGB optical frames. The pipeline also successfully isolated and classified peripheral sensors, such as the forearm cameras and the gripper-mounted accelerometers. A minor semantic limitation was observed with the $laser\_tilt\_link$, which was correctly identified as a $RangingSensor$ but labeled with a generic $SensorSubpart$ instead of a specific laser origin, likely due to the tilting mount compared to the static $base\_laser\_link$.

\paragraph{End-Effector:} The dual parallel-jaw grippers of the PR2 were successfully identified and trimmed to their respective wrist origins. The grouping task proved robust, clustering diverse sub-links (including physical finger tips, operational frames, and secondary hardware like the led frame) into two symmetric \textit{ParallelGripper} assemblies. During pre-classification, the model accurately identified the physical contact points (e.g., \textit{r\_gripper\_l\_finger\_tip\_link}) as \textit{ContactSurface}. However, mirroring the "physicality bias" seen in the TIAGo evaluation, the virtual operational frames were misclassified: the \textit{r\_gripper\_tool\_frame} was labeled as $ToolHousing$ rather than a reference frame, and the virtual center point (\textit{r\_gripper\_l\_finger\_tip\_frame}) was categorized as a generic \textit{StructuralSubpart}. Despite these semantic misalignments at the sub-part level, the kinematic kinship established during the grouping stage ensured the structural integrity of the final $ParallelGripper$ units.

\subsection{Pepper evaluation} 

The Pepper robot represents a more complex challenge for automated abstraction due to the high density of exteroceptive sensors mounted on its mobile base and the social-interaction components located on its head and hands. As a benchmark for social robotics, Pepper integrates three cameras and arrays of sonar sensors, bumpers, and capacitive touch zones, requiring the pipeline to distinguish between functionally similar sensors distributed across different spatial sectors.

\paragraph{Sensor:} The abstraction process successfully identifying 26 distinct functional units. In the mobile base, the pipeline accurately partitioned the six $Lidar$ units, two $Sonar$ frames, and three $Bumper$ components. Notably, the pipeline successfully distinguished between components that share a common root mount but serve distinct functional roles. Indeed, it correctly resisted over-merging the front and back sonar/bumper units, preserving their directional independence. While the three capacitive touch zones were correctly identified as $TactileSensor$ units with $CapacitiveSurface$ sub-parts, they remained independent entities in the final abstraction. Much like the TIAGo's microphone array, these sensors are located on a shared bodypart (the upper head) and serve a unified purpose-tactile function. This result suggests that for a more refined abstraction, it might require a semantic-spatial clustering layer that can regroup proximity-based sensors into a $TactileArray$ or $SensorArray$ component, even in the absence of explicit kinematic nesting. Furthermore, the pipeline mapped the non-standard interaction hardware $ChestButton\_frame$, as a $Bumper$ with a $ContactSurface$ sub-type. However, given its role in user-initiated signaling, a classification as a $TactileSensor$ would more accurately reflect its functional intent. Finally, in the head, the dual RGB cameras and the 3D depth sensor were correctly isolated into $MonocularCamera$ and $VisionSensor$ components respectively, accurately resolving the specific $RgbFrame$ and $DepthFrame$ sub-types.

\paragraph{End-Effector:}
The abstraction of Pepper's end-effectors successfully managed the high-dimensional branching of its five-fingered hands. Both assemblies were correctly trimmed at the wrist origins and classified as $AnthropomorphicHand$ components. This result is significant as it demonstrates the pipeline's ability to navigate multi-link kinematic subchains, such as the three-link structure of each finger, to identify the distal $ContactSurface$ segments. Interestingly, the Pepper model achieved the highest semantic accuracy for operational frames among the three robots. Unlike the TIAGo and PR2 abstractions, where tool frames were often misclassified as $ToolHousing$, the pipeline correctly identified the $l\_gripper$ and $r\_gripper$ links as $ToolCenterPoint$. This improvement likely stems from the specific naming convention used in the Pepper URDF and a more streamlined kinematic mount that avoids the bulky physical "palm" housing seen in the other platforms.

\subsection{Discussion}

The abstraction results across the TIAGo, PR2, and Pepper platforms demonstrate that kinematic chains and link identifiers provide sufficient input for automated ontological population, enabling the robust identification of component groups and their constituent sub-links while ensuring alignment with the ontology's taxonomy. A primary strength of the pipeline is its structural fidelity as by anchoring the grouping logic in the kinematic tree, the system consistently avoids the "hallucination" of non-existent connections. Moreover, even when the pre-classification stage encountered semantic ambiguity, such as labeling virtual frames as $ToolHousing$ in TIAGo and PR2, the grouping mechanism ensured these frames were correctly integrated into their respective end-effector assemblies rather than being discarded or isolated. This suggests that while LLMs may struggle with the "physicality bias" of abstract reference frames, the kinematic subchain acts as a robust fail-safe that preserves the functional integrity of the robot's hardware model.

However, the evaluation also revealed a persistent limitations inherent to "flat" URDF structures. As observed with TIAGo's microphone array and Pepper's head-touch sensors, the absence of intermediate mounting links in the URDF prevents the pipeline from automatically clustering spatially adjacent sensors into unified arrays. However, it is important to note that since the URDF is the sole source of grounded information provided to the system, this is an information-theoretic constraint rather than a failure of the pipeline logic. Without visual data or spatial proximity information, a human expert would similarly be unable to infer these groupings based on naming conventions alone. In these cases, maintaining separate instances is the most conservative and accurate interpretation of the provided structural data.
In contrast, the classification of Pepper’s $ChestButton$ as a $Bumper$ represents a different challenge: the trade-off between mechanical reasoning and interaction context. While the model correctly identified the physical nature of the component (a contact-based input), it lacked the higher-level functional context to distinguish a navigation-safety sensor from an HRI interface. To account for this, future iterations could implement a more informed closed-loop mechanism that leverages the ontology's property restrictions to disambiguate similar components. For example, while a Bumper is typically defined by its location on the robot's base for collision detection, a tactile sensor could be constrained by properties defining its role in user interaction and its placement on the torso. By encoding these ontological requirements, the system could automatically flag the $ChestButton$ as a tactile component despite its mechanical resemblance to a bumper.

Regarding the partially-closed loop mechanism embedded in the pipeline, it is important to note that the LLM is strictly constrained to the provided taxonomy. Consequently, if this taxonomy is either insufficiently detailed or semantically mismatched (failing to align with the human interpretation of robotic functions) the LLM’s output will logically reflect these misalignments (e.g., \textit{ChestButton}). Without a distinct $UserInterfaceDevice$ or $SocialInteractionComponent$ class to act as a more precise anchor, the LLM maintains ontological integrity by choosing the closest valid parent, even at the cost of functional nuance.

To mitigate these misinterpretations, we could add a more informed closed-loop mechanism which not only takes as input the ontology's taxonomy but also the specific properties that apply to each concept. For example, while a camera component usually consists of an optical frame, it also requires properties defining its field of view and focal parameters. Similarly, by encoding the property that bumper components are restricted to the navigation base, the system could flag the torso-mounted chest button as a mismatch. This would trigger the reasoning engine to seek an alternative classification, such as a \textit{TactileSensor}, which possesses properties more consistent with torso-level human-robot interaction. Such a mechanism would extend the pipeline from iterative labeling tasks to iterative process semantic verifications.
 
\section{Conclusion}

In this paper, we presented a methodology for automating ontology population from semi-structured URDF descriptions, effectively bridging the gap between raw kinematic data with structured domain-expert knowledge. By leveraging LLMs as classification heuristics rather than open-ended generators, we demonstrated that hardware abstraction can be achieved with high fidelity when grounded by formal taxonomic constraints.

Our first results confirm that knowledge-constrained loop mechanism successfully emulates the reasoning of a human expert-capable of interpreting informal naming conventions while maintaining strict alignment with the robot's existing knowledge state. This approach ensures that the resulting knowledge base remains consistent with defined taxonomies, guaranteeing that automated inference remains semantically valid. Such a bridge is a prerequisite for the targeted explainable HRI. Indeed, by transforming "links and joints" into "arms and sensors" the robot gains the structured self-awareness necessary to parse human instructions and explain its own physical limitations.

While this study focused on sensors and end-effectors, we think the methodology to be extensible to a robot's entire physical embodiment. Moving toward a full-body representation will exploit the LLM's commonsense reasoning to resolve the diverse nomenclature used across platforms—identifying, for example, that an "arm" is a functional assembly of an upper arm, elbow, and forearm regardless of specific string identifiers.

Beyond the consideration of the whole body, future research will focus on the integration of formal semantic verification prior to population. We propose using SHACL shapes to define mandatory attributes for specific components, such as requiring a camera to possess a modality type (RGB, Depth, IR) and a reference optical frame. In addition, while we only consider basic relations like $hasComponent$, future iterations will aim to automatically populate semantically richer relations such as $hasOpticalFrame$, $hasReferenceFrame$, or $isMountedOn$. This move toward relational complexity will enable a more comprehensive robotic world model, allowing the architecture to understand not just what its components are, but their precise functional roles.

\begin{credits}
\subsubsection{\ackname} 
This research was funded by the French National Research Agency (ANR) under the HumFleet project (grant number 23-CE33-0003).

\subsubsection{\discintname}
The authors have no competing interests to declare that are relevant to the content of this article.
\end{credits}

\bibliographystyle{splncs04}
\bibliography{biblio}

\end{document}